\pdfoutput=1

\documentclass[sigconf]{acmart}
\AtBeginDocument{%
  \providecommand\BibTeX{{%
    \normalfont B\kern-0.5em{\scshape i\kern-0.25em b}\kern-0.8em\TeX}}}
\setcopyright{acmcopyright}
\copyrightyear{2022}
\acmYear{2022}
\acmConference[ACM WebSci '22]{ACM WebScience}{June 26--29,
  2022}{Barcelona, Spain}
\acmISBN{978-1-4503-XXXX-X/18/06}

\usepackage{times}
\usepackage{latexsym}
\usepackage[T1]{fontenc}
\usepackage[utf8]{inputenc}
\usepackage{microtype}
\usepackage{colortbl}
\definecolor{dGray}{gray}{.6}
\definecolor{mGray}{gray}{.9}
\definecolor{lGray}{gray}{.94}

\usepackage{xparse}
\NewDocumentCommand\gui{}{\includegraphics[scale=0.1]{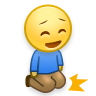}}
\NewDocumentCommand\exin{}{\includegraphics[scale=0.1]{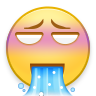}}
\NewDocumentCommand\confusedFace{}{\includegraphics[scale=0.1]{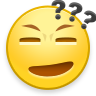}}
\NewDocumentCommand\ziya{}{\includegraphics[scale=0.1]{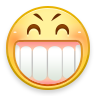}}
\NewDocumentCommand\zuiyou{}{\includegraphics[scale=0.1]{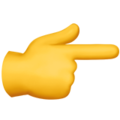}}
\NewDocumentCommand\facetears{}{\includegraphics[scale=0.08]{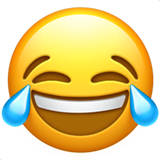}}
\NewDocumentCommand\usflag{}{\includegraphics[scale=0.08]{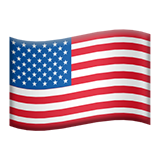}}
\NewDocumentCommand\emojiU{}{\includegraphics[scale=0.08]{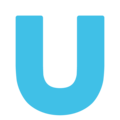}}
\NewDocumentCommand\emojiss{}{\includegraphics[scale=0.08]{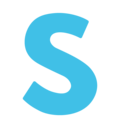}}

\usepackage{graphicx}

\title{Identifying and Characterizing Active Citizens who Refute Misinformation in Social Media}

\author{Yida Mu}
\affiliation{%
  \institution{The University of Sheffield}
  \city{Sheffield}
  \country{UK}}
   \email{y.mu@sheffield.ac.uk}

\author{Pu Niu}
\affiliation{%
  \institution{Central China Normal University}
  \city{Wuhan}
  \country{China}}
  \email{nphenu@126.com}
  
\author{Nikolaos Aletras}
\affiliation{
  \institution{The University of Sheffield}
  \city{Sheffield}
  \country{UK}}
  \email{n.aletras@sheffield.ac.uk}

\begin{document}


\begin{abstract}
The phenomenon of misinformation spreading in social media has developed a new form of active citizens who focus on tackling the problem by refuting posts that might contain misinformation. Automatically identifying and characterizing the behavior of such active citizens in social media is an important task in computational social science for complementing studies in misinformation analysis. In this paper, we study this task across different social media platforms (i.e., Twitter and Weibo) and languages (i.e., English and Chinese) for the first time. To this end, (1) we develop and make publicly available a new dataset of Weibo users mapped into one of the two categories (i.e., misinformation posters or active citizens); (2) we evaluate a battery of supervised models on our new Weibo dataset and an existing Twitter dataset which we repurpose for the task; and (3) we present an extensive analysis of the differences in language use between the two user categories.\footnote{Data is available here: \url{https://github.com/YIDAMU/Misinfo_Debunker_and_Spreader}}
\end{abstract}

\maketitle

\section{Introduction}
The diffusion of misinformation in social media has far-reaching implications on society (e.g., political polarization, election manipulation). Misinformation propagates faster than credible information among users in social media \citep{vosoughi2018spread}, whilst coming across a non-factual story once, it is enough to increase later perception of its accuracy \citep{pennycook2018prior}.

To combat misinformation, several fact-checking platforms (e.g., Snopes\footnote{\url{https://www.snopes.com/}} and the Weibo Rumour Reporting Platform\footnote{\url{http://service.account.weibo.com}}) have been created with the aim to provide evidence on why particular claims are not factually correct (i.e., debunking or fact checking). This has subsequently resulted in a new form of active citizenship with large number of social media users directly reporting suspicious posts or actively sharing posts with evidence to refute claims made by other users which are likely to contain misinformation. Other examples of active citizenship include civic engagement, political activism, community help, volunteering and neighborhood associations~\cite{johansson2007we}. In scope of social media, we consider users who actively debunk misinformation as active citizens since they work to make a difference (i.e., debunking misinformation) in online communities (e.g., social media platforms).

Automatically identifying and analyzing the behavior of active citizens in social networks is important for diffusion of misinformation prevention at the user level~\citep{singh2020multidimensional,rangel2020overview,giachanou2020role}. It can be used by (1) social media platforms (e.g., Facebook\footnote{\url{https://www.facebook.com/journalismproject/programs/third-party-fact-checking/how-it-works}}) to track suspicious posts at an early stage (e.g., reports of suspicious posts from end users); (2) psychologists to complement studies on analyzing personality traits of those users who spread or debunk unreliable posts \citep{pennycook2019lazy,pennycook2020falls}; and (3) fact-checking websites to develop personalized recommendation systems to assist active citizens in correcting suspicious posts \citep{vo2018rise,you2019attributed,karmakharm-etal-2019-journalist}.

Previous work on automatically identifying active citizens who refute misinformation has focused only on a single social media platform (i.e., Twitter) using a relatively small dataset (e.g., with only 454 users) consisting of users tweeting in English \citep{giachanou2020role}. In addition, most of these previous studies have used supervised machine learning models with features extracted from text (e.g., bag-of-words, topics, psycho-linguistic information) and task-specific neural models trained from scratch without exploring state-of-the-art pretrained large language models~\citep{devlin2019bert}.

The purpose of this paper is to study the differences in language use between the two user categories: (i) users who share suspicious posts (i.e., misinformation posters) and (ii) users who actively debunk misinformation (i.e., active citizens) To this end, we pose the following two research questions:
\begin{itemize}
\item  Can we automatically identify active citizens and misinformation posters based on their language use in social media?
\item  Can we characterize the linguistic differentiation between the two groups of users?
\end{itemize}

\noindent To answer these research questions: 

\begin{itemize}
    \item We develop a new large publicly available dataset from Weibo consisting of 48,334 users labeled either as active citizens or misinformation posters;
    \item We repurpose\footnote{The dataset has been used in computational misinformation analysis for automatic generation of fact-checking tweets and recommender systems for fact-checking \citep{vo2018rise,vo2020standing}.} an existing dataset developed by \citet{vo2019learning} to model the task of predicting active citizens and misinformation posters on Twitter; 
    \item We evaluate several state-of-the-art pretrained neural language models adapted to the task. Due to the fact that the user text can be very long (e.g., thousands of posts), we develop efficient hierarchical transformer-based networks achieving up to 85.1 and 80.2 macro F1 scores on Weibo and Twitter respectively; 
    \item We finally provide an extensive linguistic analysis to highlight the differences in language use between active citizens and misinformation posters. We also provide a qualitative analysis of the limitations of our best models in predicting accurately whether a user is an active citizen or a misinformation poster.
\end{itemize}

\section{Related Work}
\subsection{Misinformation: Definition and Types}
Misinformation in social media can be generally defined as any false or incorrect information (e.g., fabricated news, rumors, etc.) that is published and propagated by end users \citep{wu2019misinformation}. Particularly, unverified posts in social media (i.e., rumors) are defined as any item of circulated information whose veracity is yet to be verified at the time of posting \citep{zubiaga2018detection}.

\subsection{Misinformation Detection}
Previous work on computational misinformation detection has focused on predicting the credibility or bias of news articles \citep{rashkin2017truth,perez2018automatic,baly2020we} and news sources \citep{Baly2018,aker2019credibility}. 
To prevent wide spread of misinformation, propagation-based detection methods are employed to enable early misinformation detection in social media \citep{zhou2019early,tian2020early,xia2020state}. In addition to using textual information, previous work on automated fact-checking also jointly use images and user profile information extracted from metadata associated with unreliable posts \citep{lee2011seven,vo2020facts}.

Common automated fact-checking frameworks rely on external knowledge to determine the credibility of an unverified post, and they usually include one or more information retrieval models \citep{hanselowski2018ukp,nie2019combining}. Pre-trained language models (e.g., BERT \citep{devlin2019bert}) have recently been applied for fact-chekcing without using any external knowledge \citep{jobanputra2019unsupervised,lee2020language,williams2020accenture} since they encapsulate factual knowledge from the massive amount of data used for pre-training, e.g., the English Wikipedia and Books Corpus \citep{devlin2019bert}.  

These misinformation detection tasks mentioned above are usually performed on existing datasets, e.g., Liar \citep{wang2017liar} and FEVER \citep{thorne2018fever} that typically contain claims associated with a label denoting if it is factual or not. These datasets do not usually include information on the user made the claim. Based on the publicly accessible Weibo Rumor Reporting Platform, \citet{liu2015statistical} developed a Weibo rumor dataset with 7,055 misinformation posters and 4,559 active citizens, however, many users no longer exist as these rumor cases were collected between 2011 and 2013. Similarly, \citet{song2019ced} collected 3,387 rumor cases with their corresponding original publishers and 2,572,047 users who repost these fact-checked rumors.

\subsection{User Behavior Analysis Related to Misinformation}
Previous work in sociology and psychology have mostly used traditional survey-based methods to explore the personality traits \citep{pennycook2018prior,talwar2019people} and behavior \citep{altay2019so,tandoc2020diffusion} of misinformation posters. A US-based survey shows that consumers of reliable mainstream news media are more likely to use fact-checking websites for checking the factuality of news claims \citep{robertson2020uses}. Besides that, social media users are more inclined to trust debunked information that was shared by their network of friends rather than strangers \citep{margolin2018political}. 

Existing work on using computational approaches to misinformation analysis has analyzed the difference of users' reactions (e.g., reply or retweet) to unreliable news sources and mainstream media as well as their characteristics (e.g., user demographic information) \citep{glenski2018identifying,glenski2018propagation}. 
To detect malicious accounts on social media, \citet{addawood2019linguistic} and \citet{luceri2020detecting} have focused on identifying political trolls that diffuse misinformation and politically biased information during the US 2016 democratic election. \citet{mu2020identifying} and \citet{rangel2020overview} focus on identifying Twitter users who diffuse unreliable news stories either on post level or news media level. 

\citet{vo2019learning} uncover the positive impact of misinformation active citizens on preventing the spread of false news. They found that around 7\% of the original tweets (among 64k tweets) are irretrievable within five months of being debunked due to the suspension of the Twitter account and the deletion of tweets. This suggests that developing downstream tools (e.g., automatic generation of personalized fact-checking tweets \citep{vo2020facts,vo2020standing}) can encourage misinformation active citizens to actively prevent the spread of misinformation and help social media platforms to suspend malicious users. Moreover, active citizens who are active in sharing fact-checking information are found to use less informal language including swear words and are more likely to engage in debunking reliable posts about politics and fauxtography, i.e., photo edited images with misleading content~\citep{vo2018rise}. \citet{giachanou2020role} explore the impact of using linguistic features and user personality traits on identifying fake news posters and checkers based on 2,357 Twitter users. 

Our work, on the other hand, is the first attempt to model active citizens who refute misinformation across different social media platforms and languages using our newly developed Weibo dataset and a substantially larger Twitter dataset than the one used by \citet{giachanou2020role} which has not been employed yet for this task.

\section{Task and Data}
\subsection{Task Description}
Following \citet{giachanou2020role}, we frame a binary classification task aiming to distinguish between users that tend to diffuse misinformation (i.e., misinformation posters) and users who actively tend to refute such unreliable posts (i.e., active citizens) using language information. Note that one could also use a user’s social network information for modeling the predictive task but this is out of the paper’s scope because we are interested in analyzing differences in language use between the misinformation posters and active citizens across different social media platforms. Given a set of social media users, our task is to train a supervised classifier that can learn relations between users' linguistic patterns (i.e., the collection of users' original posts) and the corresponding class (i.e., misinformation posters or active citizens).

\begin{figure}[t!]
    \centering
    \includegraphics[scale=0.3]{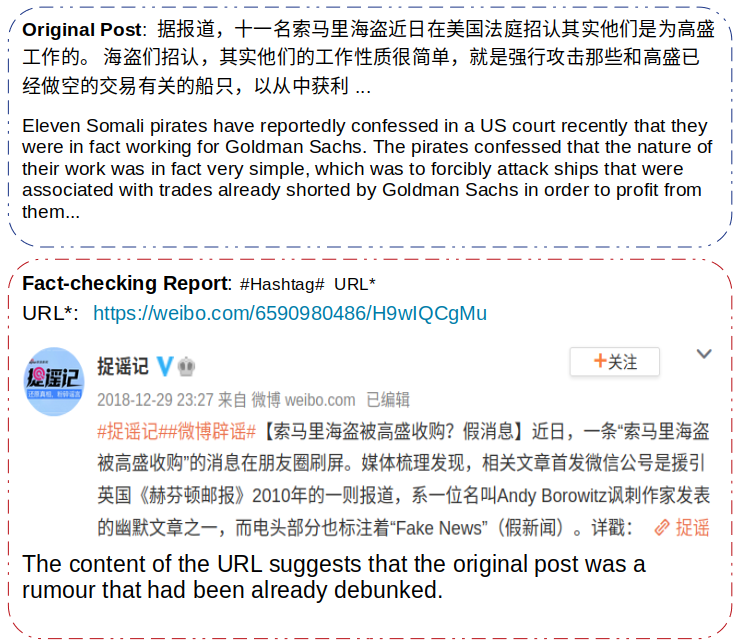}
    \caption{An example of a pair of misinformation poster (in the blue box) and debunker on Weibo with the corresponding fact-checking information (in the red box)}
    \label{fig2}
\end{figure}

\begin{figure}
    \centering
    \includegraphics[scale=0.3]{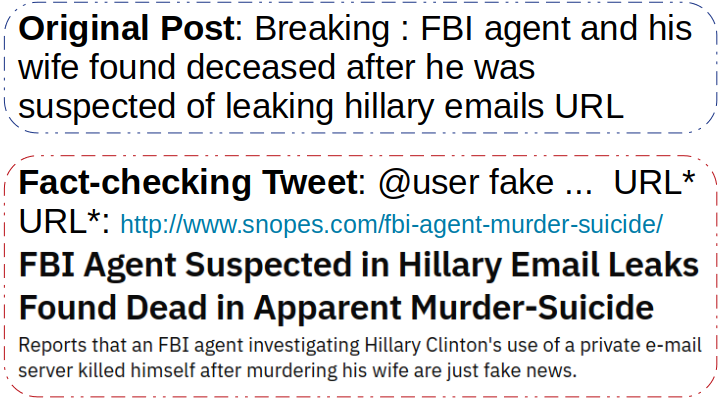}
    \caption{An example of a pair of misinformation poster and debunker on Twitter with the corresponding fact-checking information}
    \label{fig1}
\end{figure}

\subsection{Weibo Data}
In 2012, Weibo developed an official community management center to receive reports from end users for all kinds of malicious posts including misinformation, hate speech and content plagiarism. To combat misinformation, Weibo provides a fact-checking platform for its users to report any suspicious misinformation posts which are then subsequently fact-checked officially by the platform. These Weibo posts are eventually labeled as \emph{true} or \emph{false}. Alternatively, a post may also remain \emph{unverified} until it has been fact-checked. In case that a post has been deemed to be labeled as \emph{false}, it is also accompanied with debunking information refuting the claim. Figure~\ref{fig2} shows an example including the original publisher (e.g., misinformation poster), active citizen and fact-checking information on Weibo. 

\paragraph{Collecting misinformation posters and active citizens} For the purpose of our experiments, we collect 38,712 debunked cases (between 2012 and 2020) from the official Weibo Community Management Center\footnote{\url{https://service.account.weibo.com/?type=5&status=4}} following a similar data collection approach as previous work~\citep{wu2015false,ma2016detecting}. We keep only those posts that have been judged as \emph{false} and then collect the corresponding poster (i.e., the original user that has published the post) and up to 20 recent active citizens (i.e., users that have officially reported that the post contains misinformation). Note that the official Weibo platform only allows access to the earliest 20 active citizens, even though some suspicious posts have been reported and refuted by more than 20 users. However, we notice that only in 709 cases there are more than 20 active citizens (i.e., less than 2\% of all debunked posts). Given that some user accounts may have been suspended or become private, we remove those from the data. 

Note that there is a difference in the definition of active citizens in Weibo and Twitter datasets. In Weibo, all active citizens are those who report misinformation to the Weibo Official Fact-checking Platform. In Twitter, active citizens are defined as ones who cite fact-checking URLs to refute misinformation. For consistency, we label all of these users as active citizens since they both \textbf{actively} try to refute misinformation. Weibo active citizens are not required to provide evidence or fact-checking ULRs but they are free to report a post on a suspicion that it contains misinformation.

\paragraph{Collecting User Posts} We use the Weibo API\footnote{\url{https://open.weibo.com/development/businessdata}} to collect up to 2,000 posts for each user since the median number of user posts are 968 and 855 for the two user categories (i.e., poster and debunker) respectively. We only consider users with more than 30 original posts and filter out all users who have both spread and debunk misinformation posts. After removing duplicate users, the final dataset contains 22,632 distinct posters and 25,702 distinct active citizens respectively.

\begin{table}[!t]
\caption{Data Statistics.}
\centering
\small
\begin{tabular}{|l|l|l|l|l|}
\hline
 \rowcolor{dGray}                 & \multicolumn{2}{l|}{\bf Weibo} & \multicolumn{2}{l|}{\bf Twitter}     \\ \hline
\rowcolor{mGray} {\bf \#Users}                  & {\bf Poster}      & {\bf AC}    & {\bf Poster}    & {\bf AC} \\ \hline
                  & 22,632           & 25,702          & 15,696          & 17,293         \\ \hline
\rowcolor{mGray} {\bf \#Posts}  & \multicolumn{4}{l|}{}                                        \\ \hline
Min               & 31            & 31           &30            &  30         \\ \hline
Max               & 2,000         & 2,000        &3,200         &  3,200      \\ \hline
Mean              & 596           & 576          &2,932         &  2,824      \\ \hline
Total             & 13.5M         & 14.8M        &46.0M         &  48.8M      \\ \hline
\rowcolor{mGray} {\bf\#Tokens/User} &               &              &              &            \\ \hline
Min               & 127           &126           &663           &  674        \\ \hline
Max               & 104,947       &104,801       &81,052        &  81,028     \\ \hline
Mean              & 13,643        &10,127        &33,759        &  35,652     \\ \hline
Median            & 4705          &3,730         &32,726        &  35,312     \\ \hline
\end{tabular}
\label{t:Data}
\end{table}

\subsection{Twitter Data}
\paragraph{Collecting misinformation posters and active citizens} To label Twitter users as misinformation posters or active citizens, we use a publicly available dataset with totally 73,203 users provided by \citet{vo2019learning}. 

\citet{vo2019learning} first use the Hoaxy System \citep{shao2016hoaxy} to collect fact-checking tweets (FC-tweets) that contain links to relevant fact-checking information from PolitiFact and Snopes. These FC-tweets contain users who post URLs from fact-checking websites as credible evidence to refute misinformation posts in public conversations on Twitter (i.e., active citizens). They also contain the original users whose posts are debunked (i.e., misinformation posters). Figure~\ref{fig1} shows an example that contains the original post, fact-checking tweets and the corresponding debunking information from Snopes. According to \citet{vo2019learning}, this dataset only contains misinformation active citizens who post English tweets with corresponding URLs linking to evidence (e.g., news article) that refutes a false claim.
During data exploration, we observe that some Twitter users refer the fact-checking URLs to support the personal claims of the original posters, i.e., the original message has been proven correct. Therefore, we only consider users who share fact-checking URLs to refute tweets containing misinformation, i.e. those who are flagged as \emph{False} by the corresponding fact-checking platform. In this way, we ensure that the selected active citizens have a clear intention to refute misinformation.

\paragraph{Collecting User Posts} For each Twitter user, we use the Twitter Public API\footnote{\url{https://developer.twitter.com/en/docs}} to collect up to 3,200 tweets due to limits excluding any retweets. Moreover, we filter out users with less than 30 original tweets, users that may appear in both groups and keep users with a majority of English tweets (e.g., tweets that are labeled as `en' or `en-gb' by Twitter). As in the Weibo dataset, we also remove all users that both spread and debunk misinformation since we currently focus on the binary setting as in \citet{giachanou2020role} since we found that less than 10\% of all users fall into this category (i.e., both spread and debunk misinformation) in the two datasets.\footnote{We leave this multi-label classification task for future work} This process yielded 15,696 posters and 17,293 active citizens respectively. This is approximately 100 and 15 times larger than the datasets used in prior work~\citep{giachanou2020role,rangel2020overview}.

\begin{figure*}[!t]
    \centering
    \includegraphics[scale=0.4]{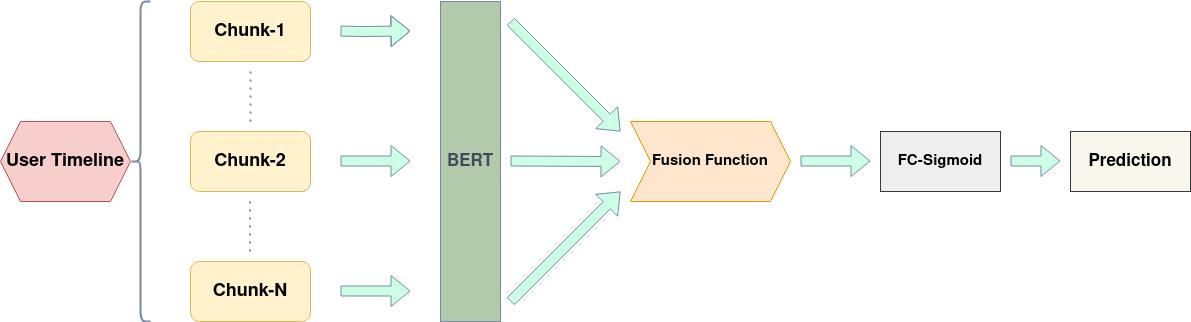}
    \caption{Overview of the hierarchical transformer architecture used in our work. N = L / 510, where N denotes the number of chunks and L denotes the number of tokens. The Fusion Function denotes how we fuse the chunk-level information into the global representation, i.e., using Max Pooling, Mean Pooling and LSTM-Attention.}
    \label{model}
\end{figure*}

\subsection{Data Statistics and Topical Coverage}
Table~\ref{t:Data} shows the descriptive statistics of the data including the number of tweets and tokens obtained from Weibo and Twitter. Both datasets cover a broader range of topics rather than a narrow subset of political misinformation. The Twitter dataset provided by \citet{vo2019learning} is developed based on two popular fact-checking platforms (i.e., Snopes and PolitiFact) and covers more topics (e.g.,  medical and business) other than politics. However, we also notice that more than 50\% of cases are related to politics given that one of the fact-checking platform (i.e., PolitiFact) is totally political oriented. As for the Weibo dataset, we collect all available cases of misinformation from the Weibo fact-checking platform from 2012 to 2020. According to \citet{liu2015statistical}, more than 90\% of debunked Weibo misinformation are related to politics, economics,  pseudo-science and social life. 

\subsection{Text Pre-processing}
\paragraph{Weibo Posts} We pre-process the Weibo raw posts by converting all traditional Chinese words to simplified Chinese and then tokenizing using JIEBA, a Chinese text processing toolkit.\footnote{\url{https://github.com/fxsjy/jieba}} We keep all non-Chinese words since we notice some common English words (e.g., \emph{python, hello, and world}) are including in the training corpus of both Chinese and English transformer models.
\paragraph{Tweets} For the Twitter data, we follow a similar pre-processing pipeline\footnote{\url{https://github.com/VinAIResearch/BERTweet}} as in \citet{nguyen2020bertweet}. In brief, we pre-process the tweets by first lowercasing and then tokenizing using the \emph{TweetTokenizer} from NLTK toolkit \citep{bird2009natural}. Besides, we further normalize each Tweet by replacing each emoji,\footnote{We use the emoji Python package \url{https://pypi.org/project/emoji/}} URL and @-mention with special tokens, i.e., \emph{single word token}, \emph{@USER} and \emph{HTTPURL} respectively.

\section{Predictive Models}
\subsection{Baseline Models}
\paragraph{Logistic Regression}
We apply logistic regression with L2 regularization penalty using Bag-of-Words (BOW) to represent each user as a TF-IDF weighted vector over a 10,000 sized vocabulary. We only keep n-grams appearing in more than 5 times and no more than 40\% of the total users. 
We also represent each user over a distribution of manually created lexical categories represented by lists of words provided by the Linguistic Inquiry and Word Count (LIWC) 2015 dictionary \citep{pennebaker2001linguistic}. LIWC has been extensively used in psycho-linguistic studies. 
\paragraph{BiLSTM-ATT} 
Furthermore, we train a  Bidirectional Long Short Term Memory network \citep{hochreiter1997long} with self-attention (BiLSTM-ATT) from scratch. The BiLSTM-ATT takes as input the users' historical posts, maps their words to pre-trained word embeddings and subsequently passes them through a bidirectional LSTM layer. A user embedding is computed as the sum of the resulting context-aware embeddings weighted by the self-attention scores. The user embedding is then passed to the linear prediction layer with sigmoid activation.

\subsection{English Transformers}
\paragraph{BERT} Bidirectional Encoder Representations from Transformers \citep{devlin2019bert} is a masked language model using a Transformer Network \citep{Vaswani2017} pre-trained on the BooksCorpus and English  Wikipedia. 

\paragraph{RoBERTa} The Robustly Optimized BERT Pretraining Approach (RoBERTa) \citep{liu2019roberta} is a BERT-style language model trained with fine-tuned hyper-parameters, larger batch size, and longer sequence compared to the original BERT. RoBERTa is trained on a combination of the original corpus used to train BERT and extra texts including English new articles and web content \citep{liu2019roberta}. 

\paragraph{Longformer} The Long-Document Transformer (Longformer) \citep{beltagy2020longformer} is pretrained using the original RoBERTa checkpoint \citep{liu2019roberta} with a sliding window attention pattern (same window size of 512 as RoBERTa) and extra positional embeddings to support a maximum length of 4,096 (from 512). Longformer can handle a longer text sequences and achieves state-of-the-art performance on long document downstream NLP tasks~\citep{beltagy2020longformer}. 

\subsection{Chinese Transformers}
\paragraph{CBERT} For our Weibo prediction task, we first employ a Chinese BERT (CBERT) \citep{cui2019pre} model pretrained using a whole word masking strategy. CBERT is trained from the existing checkpoint of the Bert-Base-Chinese\footnote{\url{https://storage.googleapis.com/bert_models/2018_11_03/chinese_L-12_H-768_A-12.zip}} model, which has the same structure (e.g., layers and parameters) as the original BERT. 

\paragraph{ERNIE} Enhanced Representation through Knowledge Integration (ERNIE) \citep{sun2019ernie} is designed to learn language representations using knowledge masking strategies, i.e., entity-level masking and phrase-level masking. ERNIE is trained on both formal (e.g., Baidu Baike, a platform similar to Wikipedia and Chinese news articles) and informal (e.g., posts from Tieba, an open discussion forum similar to Reddit) Chinese corpora.

\subsection{Handling Long Text}
Transformer-based models cannot handle long sequences in a single standard GPU card due to the large memory requirements. To deal with this issue in our datasets, we experiment with truncated and hierarchical methods for all transformer-based models.

\paragraph{Truncated Transformers}
Following similar work on modeling long texts by \citet{sun2019fine}, we first employ a simple truncation method that cuts off the input to the maximum length supported by BERT and Longformer (e.g. 512 and 4,096 tokens). Following the same strategy as in \citet{devlin2019bert}, we fine-tune transformer-based models by adding a linear prediction layer on the model special classification token, e.g., [CLS] of BERT and $<$s$>$ of Longformer respectively. 

\paragraph{Hierarchical Transformers}
Given that the majority of users' concatenated posts contain more than 512 tokens, we also use a hierarchical transformer structure \citep{pappagari2019hierarchical,mulyar2019phenotyping} (see Figure~\ref{model}) for our long document classification task. 
We first split the input sequence (i.e., the collection of users' original posts) into N = L / 510 chunks\footnote{N denotes the number of chunks and L the number of tokens.} of a fixed length e.g., 512 including task special tokens (e.g., [CLS] tokens for BERT) and 4096 tokens for Longformer. 
For each of these word chunks, we obtain the representation of the [CLS] token from the fine-tuned BERT on our dataset. We then stack these segment-level representations into a sequence, which serves as input to a LSTM layer with a self-attention mechanism to learn a user-level representation. Finally, we add two fully connected layers with ReLU and sigmoid activations respectively on top of LSTM layer as in \citet{pappagari2019hierarchical}. 

Following \citet{sun2019fine}, we also test two simple hierarchical methods by directly using max pooling and mean pooling to stack the [CLS] embeddings of all the chunks of each user into a document-level representation.

\section{Experimental Setup}
\subsection{Hyper-parameters}
For both Twitter and Weibo datasets, we train the models on the training set (70\%) and tune the hyper-parameters on the validation set (10\%). We tune the regularization parameter $\alpha \in$ \{1e-1, 1e-2, 1e-3, 1e-4, 1e-5\} of the Logistic Regression, setting $\alpha=1e-4$. 
For BiLSTM-ATT, we use 200-dimensional GloVe embeddings \citep{pennington2014glove} pre-trained on 2-billion tweets and 300-dimensional Chinese Word Vectors\footnote{\url{https://github.com/Embedding/Chinese-Word-Vectors}} \citep{li2018analogical} pre-trained on Weibo data. We tune the LSTM {\it hidden unit size} $\in$ \{50, 100, 150\} and {\it dropout rate} $\in$ \{0.2, 0.5\} observing that 150 and 0.5 perform best respectively. 
For transformer-based models, we use BERT-Base-Uncased, RoBERTa-Base and Longformer-Base-4096 models fine-tuning them with learning rate $lr \in$ \{5e-5, 3e-5, and 2e-5\} as recommended in \citet{devlin2019bert}, setting $lr=2e-5$. For the Chinese language models, we use Chinese-BERT-WWM-EXT and ERNIE-1.0 models fine-tuning them with learning rate $lr \in$\{5e-5, 3e-5, and 2e-5\} as in \citet{cui2019pre}, setting $lr=2e-5$. The maximum sequence length is set to 512 (including task special tokens, e.g., [CLS]) except the Longformer-Base-4096 which can handle a 4,096 input sequence length. 

We use a batch size of 16 for all transformer-based models except the Longformer where we use batch size of 4. During training of the neural models, we use early stopping based on the validation loss and then use the saved checkpoint to compute the model predictive performance on the test set.

\subsection{Implementation Details}
We perform all the experiments on a single NVIDIA V100 graphics card. We use the implementation of transformer-based models available from the HuggingFace library \citep{wolf2019huggingface}.

\subsection{Evaluation Metrics}
We run each model with the best hyper-parameter combination three times on the heldout set (20\%) using different random seeds, and report the averaged macro precision, recall and F1 score (mean $\pm$ standard deviation).

\section{Results}

\begin{table}[!t]
\caption{Weibo binary classification results (mean $\pm$ standard deviation). $\ddagger$ denotes that the HierERNIE LSTM performs significantly better than truncated ERNIE (t-test; $p<.05$)}
\small
\centering
\begin{tabular}{|l|c|c|c|}
\hline
\rowcolor{dGray} {\bf Model}     & {\bf P (macro)} & {\bf R (macro)} & {\bf F1 (macro)} \\ \hline

\hline
\rowcolor{mGray} {\bf Baseline Models}& \multicolumn{3}{|c|}{}\\\hline
\hspace{0.5em} LR-BOW &     $ 82.8 \pm 0.1 $ &  $ 82.7 \pm 0.1 $ & $ 82.7 \pm 0.1 $  \\
\hspace{0.5em} LR-LIWC &    $ 75.7 \pm 0.1 $ &  $ 73.7 \pm 0.2 $ & $ 74.1 \pm 0.3 $  \\
\hspace{0.5em} BiLSTM-ATT & $ 83.1 \pm 0.2 $ &  $ 82.9 \pm 0.1 $ & $ 82.9 \pm 0.1 $  \\ 
\hline
\rowcolor{mGray} {\bf Truncated Transformers} & \multicolumn{3}{|c|}{}\\\hline
\hspace{0.5em} CBERT    & $79.9\pm 0.1$ &  $ 79.9 \pm 0.2 $ & $ 79.9 \pm 0.1$ \\ 
\hspace{0.5em} ERNIE    & $80.4 \pm 0.1$ &  $ 80.5 \pm 0.2 $ & $ 80.3 \pm 0.1$  \\
\hline
\rowcolor{mGray} {\bf Hierarchical Transformers} & \multicolumn{3}{|c|}{}\\\hline
\hspace{0.5em} HierCBERT Max Pool   &$ 83.3\pm 0.2$ &  $ 83.2 \pm 0.2$ & $ 83.2 \pm 0.3$  \\
\hspace{0.5em} HierCBERT Mean Pool    &$ 84.0\pm 0.1$ &  $ 84.0 \pm 0.1$ & $ 84.0 \pm 0.1$  \\
\hspace{0.5em} HierCBERT LSTM           &$ 84.5\pm 0.1$ &  $ 84.2 \pm 0.2$ & $ 84.3 \pm 0.1$  \\
\hspace{0.5em} HierERNIE Max Pool   &$ 83.8\pm 0.1$ &  $ 83.7 \pm 0.2$ & $ 83.7 \pm 0.2$  \\
\hspace{0.5em} HierERNIE Mean Pool    &$ 84.2\pm 0.2$ &  $ 84.3 \pm 0.2$ & $ 84.2 \pm 0.2$  \\
\hspace{0.5em} HierERNIE LSTM $\ddagger$           &$ \textbf{85.2}\pm 0.1$ &  $ \textbf{85.1} \pm 0.1$ & $ \textbf{85.1} \pm 0.1$  \\
\hline
\end{tabular}

\label{t:weibo}
\end{table}

\subsection{Predictive Performance}
Tables~\ref{t:weibo} and \ref{t:TWitter} show the results obtained by all models in the Weibo and Twitter datasets.

In Twitter, HierLongformer LSTM achieves the highest F1 score overall (80.2) surpassing all the baseline models as well as the simpler hierarchical architectures, e.g., using mean and max pooling.  
For each of the transformer-based model, we observe that the hierarchical transformer architectures (e.g, LSTM, max pooling and mean pooling) outperform the truncated models across all metrics. Their hierarchical structure allows them to exploit all the available textual information from each user that impacts performance. 
The Longformer model that supports longer input sequences achieves better predictive results than the other transformer models that support shorter input sequences (e.g., BERT and RoBERTa). This is similar to results obtained by \citet{beltagy2020longformer,gutierrez2020document} where the Longformer consistently outperforms other BERT-style models in long document classification tasks.

In Weibo, HierERNIE LSTM achieves the highest F1 score overall surpassing all other models. In addition, we observe two baseline models (LR-BOW and BiLSTM-ATT) achieve a slightly lower performance than the hierarchical transformers e.g., 82.7 and 82.9 F1-score respectively. This suggests that the relationship between users' language use and labels can be learned more efficiently by using a simple classifier (e.g., LR) that has access to all users' posts, compared to a more complex model that does not use all available information. We also observe that, in general, the use of different hierarchical methods (especially the LSTM takes into account the sequence order) improve the performance of truncated transformer models. This suggests that the order of the posts and their dependencies matter. 

Lastly, we observe that the models with similar structure and characteristics trained on Weibo data are on average more accurate than the Twitter data (approximately 5\%). This highlights that input language (i.e., Chinese vs. English) and its peculiarities play an important role in the performance of text classification models.

\begin{table}[!t]
\caption{Twitter binary classification results (mean $\pm$ standard deviation). $\ddagger$ denotes that the HierLongformer LSTM performs significantly better than truncated Longformer (t-test; $p<.05$).}
\small
\centering
\begin{tabular}{|l|c|c|c|}
\hline
\rowcolor{dGray} {\bf Model}     & {\bf P (macro)} & {\bf R (macro)} & {\bf F1 (macro)} \\ \hline

\rowcolor{mGray} {\bf Baseline Models}& \multicolumn{3}{|c|}{}\\\hline
 LR-BOW           & $ 75.8  \pm0.1 $ &  $ 74.9  \pm 0.1 $ & $ 74.9 \pm 0.1$  \\
 LR-LIWC           & $68.4   \pm 0.1 $ &  $  67.4 \pm 0.1 $ & $ 67.4 \pm 0.2$  \\
 BiLSTM-ATT       & $ 76.0  \pm0.2 $ &  $ 75.0  \pm 0.2 $ & $ 75.1 \pm 0.2$  \\ 
\hline
\rowcolor{mGray} {\bf Truncated Transformers} & \multicolumn{3}{|c|}{}\\\hline
 BERT             & $73.1 \pm 0.1 $ &  $72.5 \pm 0.2 $ & $ 72.5 \pm 0.2$  \\ 
 RoBERTa          & $74.5 \pm 0.3 $ &  $73.7 \pm 0.2 $ & $ 73.7 \pm 0.2$  \\
 LongFormer       & $77.9 \pm 0.1$  &  $77.0 \pm 0.2 $ & $ 77.0 \pm 0.2$  \\
\hline
\rowcolor{mGray} {\bf Hierarchical Transformers} & \multicolumn{3}{|c|}{}\\\hline
 HierBERT Mean Pool &  $ 77.7\pm 0.2$ &  $77.4 \pm 0.2$ & $77.5 \pm 0.2$  \\
 HierBERT Max Pool &   $ 73.5 \pm 0.2$ &  $72.4 \pm 0.3$ & $ 73.6\pm 0.3$  \\
 HierBERT LSTM &          $ 78.1\pm 0.2$ &  $ 77.5\pm 0.1$ & $ 77.6\pm 0.2$  \\

 HierRoBERTa Mean Pool & $ 79.1\pm 0.3$ &  $ 78.1\pm 0.2$ & $ 78.2\pm 0.2$  \\
 HierRoBERTa Max Pool  & $ 77.5\pm 0.3$ &  $ 75.7\pm 0.3$ & $ 75.9\pm 0.3$  \\
 HierRoBERTa LSTM         & $ 78.9\pm 0.2$ &  $ 78.5\pm 0.3$ & $ 78.8\pm 0.2$  \\

 HierLongformer Mean Pool  & $ 80.3 \pm 0.2$ &  $ 79.8 \pm 0.1 $ & $ 79.9 \pm 0.1 $  \\
 HierLongformer Max Pool   & $ 79.3 \pm 0.1$ &  $ 79.0 \pm 0.1 $ & $ 79.0 \pm 0.1 $ \\
 HierLongformer LSTM $\ddagger$     & $ \textbf{80.5} \pm 0.1$ &  $ \textbf{80.1} \pm 0.1 $ & $ \textbf{80.2} \pm 0.1 $  \\
\hline
\end{tabular}

\label{t:TWitter}
\end{table}

\subsection{Model Explainability}
For both datasets, we analyze the most important input tokens that contribute to the model prediction (i.e., HierERNIE LSTM in Weibo and HierLongformer LSTM in Twitter) by employing a widely used gradient-based explainability method i.e., the \emph{InputXGrad with L2 Norm Aggregation} \citep{kindermans2016investigating} that has been found to provide faithful explanations for transformer-based models in NLP tasks~\citep{chrysostomou-aletras-2021-improving,chrysostomou2022}. The \emph{InputXGrad ($\mathbf{x} \nabla \mathbf{x}$)} ranks the input tokens  by computing the derivative of the input with respect to the model predicted class and then multiplied by the input itself, where $\nabla x_i = \frac{\partial \hat{y}}{\partial x_i}$. We then get the L2 normalized aggregation of the scores across the embedding dimensions similar to \citet{chrysostomou-aletras-2021-improving}.

\paragraph{Twitter} In Twitter, the InputXGrad scores indicate that some hashtags and emojis (note that we have detokenized wordpieces when calculating importance scores) have a higher impact on model predictions. For example, politics-related terms and hashtags (e.g., \usflag{}{}, \emph{\#POTUS}) play an important role when the model predicts Twitter users as misinformation posters. This is similar to the result from \citet{addawood2019linguistic}, showing that Twitter users using a higher number of political hashtags are more likely to be identified by the model as political trolls.
On the other hand, some tokens related to daily activities (e.g., \emph{\#yoga, \#marchforscience, \#vegetarian}) and social issues (e.g., \emph{\#blacklivematters}) are more prevalent in misinformation active citizens.

\paragraph{Weibo} In Weibo, when the model predicts users as misinformation posters, some tokens that express emotions ((e.g., \emph{surprise, unhappy and amazing})) become the key factors. In contrast, model assigns importance to some popular buzzwords (e.g., \emph{hahahaha, xswl (i.e., LMAO in Chinese abbreviations}) and celebrities (e.g., \emph{tfboys, uzi, and blackpink}) when it predicts users as misinformation active citizens. These users who tend to debunk misinformation appear to be common users using Weibo for social interactions with friends.\footnote{The letters U and S, which can be used as part of a regional indicator pair to create emoji flags for various countries.}

\begin{table}[!t]
\caption{N-grams associated with Twitter misinformation posters and active citizens sorted by Pearson's correlation ($r$) between the normalized frequency and the labels ($p < .001$).}
\centering
\small
\begin{tabular}{|l|l||l|l|}
\hline
\rowcolor{dGray} \multicolumn{4}{|c|}{{\bf n-grams}} \\ \hline
\rowcolor{mGray} {\bf Posters}            & {\bf r}         & {\bf Active Citizens}           & {\bf r}       \\ \hline
illegals        & 0.173                      & slightly      & 0.154 \\ \hline
msm (mainstream media)             & 0.165                      & empathy       & 0.154 \\ \hline
\emojiU{}{} (regional indicator)      & 0.158                      & theories      & 0.144 \\ \hline
\emojiss{}{} (regional indicator)     & 0.154                      & generally     & 0.141 \\ \hline
soros           & 0.143                      & equivalent    & 0.137 \\ \hline
brennan         & 0.142                      & necessarily   & 0.135 \\ \hline
communist       & 0.140                      & confusing     & 0.135 \\ \hline
schumer         & 0.139                      & fewer         & 0.131 \\ \hline
leftist         & 0.130                      & quotes        & 0.129 \\ \hline
rino            & 0.128                      & actively      & 0.129 \\ \hline            
\end{tabular}

\label{t:bowtwitter}
\end{table}

\begin{table}[!t]
\caption{N-grams associated with Weibo misinformation posters and active citizens sorted by Pearson's correlation ($r$) between the normalized frequency and the labels ($p < .001$). We translate all the Chinese N-grams into English.}
\centering
\small
\begin{tabular}{|l|l||l|l|}
\hline
\rowcolor{dGray} \multicolumn{4}{|c|}{{\bf n-grams}} \\ \hline
\rowcolor{mGray} {\bf Posters}             & {\bf r}         & {\bf Active Citizens}           & {\bf r}       \\ \hline
cherish          & 0.177      &  WTF           & 0.256  \\ \hline
understand       & 0.172      &  LMAO          & 0.249  \\ \hline
present          & 0.171      &  \zuiyou{}{}    & 0.239  \\ \hline
this morning     & 0.154      &  \gui{}{}       & 0.232  \\ \hline
because          & 0.149      &  rightmost    & 0.225  \\ \hline
contact          & 0.148      &  \confusedFace  & 0.222  \\ \hline
rose             & 0.145      &  awesome       & 0.218  \\ \hline
strong           & 0.143      &  Ahhhh         & 0.214  \\ \hline
\ziya{}{}   & 0.142  &  \exin{}{}    & 0.202  \\ \hline
creation         & 0.139      &  F*ck          & 0.200  \\ \hline 
\end{tabular}

\label{t:bowweibo}
\end{table}

\subsection{Error Analysis}
We also perform an error analysis by inspecting cases of wrong predictions in both datasets. We first observe that Twitter active citizens who are wrongly classified as posters are more prevalent in posting about politicians (e.g., \emph{Obama, Clinton} and \emph{POTUS}) and some hashtags (e.g, \emph{\#VOTEBIDEN, \#BIDEN2020}) related to the democratic party in the U.S.. These users are misclassified by the model possibly due to similar language use with those spreading misinformation. 
We also notice that Weibo misinformation posters who are misclassified as active citizens use cyber slang while are also more likely to express emotions e.g., \emph{Ahhh} and \facetears{}{}. We finally observe that a higher proportion of Weibo misinformation posters who are wrongly classified as active citizens are verified users (15\%) (note that higher percentage of posters (21\%) are verified users than active citizens (10\%)).

\section{Linguistic Analysis}
We further perform a linguistic analysis to uncover the differences in language use between users in the two categories, i.e. misinformation posters and active citizens. To that end, we employ univariate Pearson’s correlation test to characterize which linguistic features (i.e., BOW and LIWC\footnote{We use the LIWC English \citep{pennebaker2001linguistic} and Simplified Chinese \citep{huang2012development} dictionaries.}) are high correlated with each class following \cite{Schwartz2013}. This approach has been widely used in similar NLP studies~\citep{preotiuc-pietro-etal-2019-automatically,maronikolakis-etal-2020-analyzing,mali2022}.

\begin{figure}[t!]
    \centering
    \includegraphics[scale=0.3]{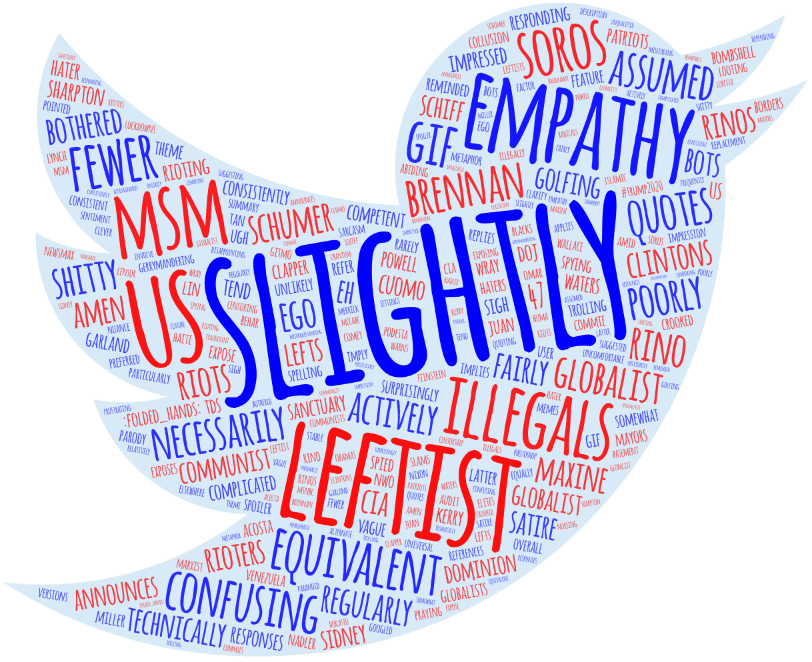}
    \caption{Words associated with misinformation posters (Red) and active citizens (Blue) on Twitter.}
    \label{fig3}
\end{figure}

\begin{figure}[t!]
    \centering
    \includegraphics[scale=0.3]{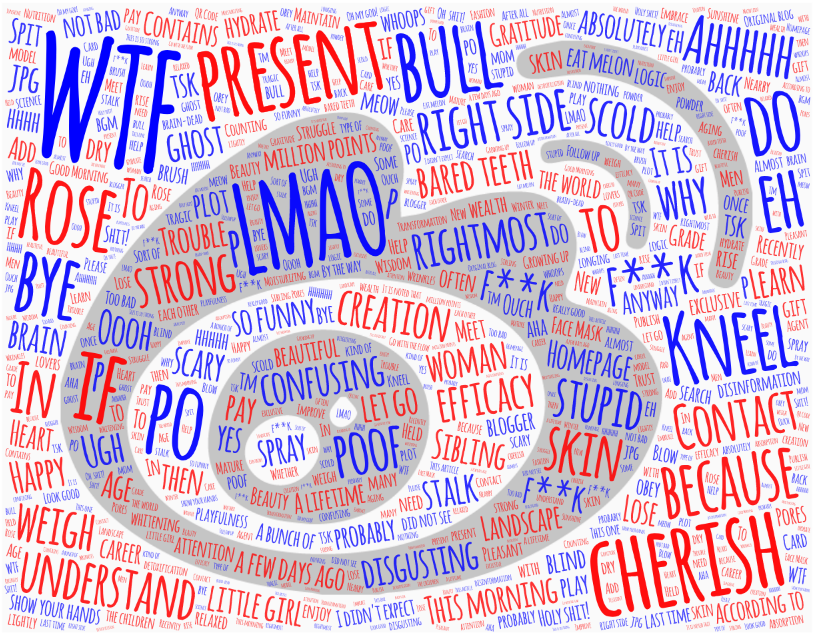}
    \caption{Words associated with misinformation posters (Red) and active citizens (Blue) on Weibo. We translate all the Chinese words into English.}
    \label{fig4}
\end{figure}

\subsection{N-grams}
Table~\ref{t:bowtwitter} shows that Twitter users who diffuse misinformation are more prevalent in posting about politics (e.g., \emph{\emojiU{}{}\emojiss{}{}, Soros} and \emph{Brennan}). This is similar to findings by \citet{mu2020identifying}, which showed that people who often retweeted news items from unreliable news sources (e.g., Infowars, Disclose.tv) are more likely to discuss politics. Moreover, active citizens on Twitter use more frequently adverbs (e.g., \emph{slightly, generally, and necessarily} and words that denote uncertainty (e.g., \emph{confusing}). 
Table~\ref{t:bowweibo} shows that Weibo active citizens are more likely to use words related to self-disclosure, e.g., \emph{WTF, LMAO} and \emph{awesome} and net-speak words e.g., \zuiyou{}{} and \emph{rightmost}. These buzzwords are more popular among average Weibo users who share interesting posts with their friends or reply to something entertaining. Note that most of Weibo active citizens are not official accounts (i.e., unverified users) which rarely use these words. Similarly, Weibo active citizens also use emojis that express uncertainty, e.g., \confusedFace{}{}, more frequently. We finally observe that Weibo misinformation posters use causation words (e.g., because). This is different from earlier studies that found deceivers to use a smaller number of causation words when telling false stories \citep{hancock2007lying}.

To better highlight the similarities between words associated with misinformation posters and active citizens, we also create two word clouds to display up to 100 N-grams features per user category in Twitter (see Figure~\ref{fig3}) and Weibo (see Figure~\ref{fig4}) datasets (i.e., the larger the font, the higher the Pearson correlation value). 

We observe that the active citizens on both platforms like to use adverbs, for example, to express certainty (e.g., absolutely, equally, and particularly). Moreover, certain words are found to be strong indicators of truthfulness according to the interpersonal deception theory \citep{buller1996interpersonal,addawood2019linguistic}. In addition, active citizens also use words (e.g., disinformation, misleading, parody, and satire) which are partly used in fact-checking tweets to debunk suspicious posts on Twitter \citep{vo2018rise}.
Compared to Weibo, Twitter misinformation posters discuss more global political events (i.e., politicians and parties) since Twitter is an international social media platform, while Weibo is used primarily by Chinese speakers.

\subsection{LIWC}

\begin{table}[!t]
\caption{English LIWC features associated with Twitter misinformation posters and active citizens sorted by Pearson's correlation ($r$) between the normalized frequency and the labels ($p < .001$).}
\centering
\small
\begin{tabular}{|l|l||l|l|}
\hline
\rowcolor{dGray} \multicolumn{4}{|c|}{{\bf LIWC}} \\ \hline
\rowcolor{mGray} {\bf Posters}            & {\bf r}         & {\bf Active Citizens}           & {\bf r}       \\ \hline
power   &0.201      &tentat     & 0.250 \\ \hline
drives  &0.186      &differ     & 0.217 \\ \hline
colon   &0.133      &adverb     & 0.201 \\ \hline
clout   &0.132      &cogproc    & 0.200 \\ \hline
we      &0.132      &insight    & 0.194 \\ \hline
exclam  &0.126      &ipron      & 0.182 \\ \hline
otherP  &0.125      &conj       & 0.181 \\ \hline
female  &0.115      &compare    & 0.174 \\ \hline
relig   &0.114      &function   & 0.160 \\ \hline
affiliation & 0.113 &comma      & 0.151 \\ \hline       
\end{tabular}
\label{t:liwcweibo}
\end{table}

Tables~\ref{t:liwcweibo} and \ref{t:liwctwitter} show the ten most correlated LIWC categories with each user class in Weibo and Twitter datasets respectively. 
We observe that users who belong to the misinformation poster class in both social media platforms are more prevalent in posting topics about \emph{Biological Processes} (e.g., \emph{female}, \emph{sexual} and \emph{health}) and \emph{Core Drives and Needs} (e.g., \emph{Power}, \emph{Drives}, \emph{Affiliation} and \emph{Achievement}). Users who refute misinformation on social network post topics related to \emph{Cognitive Processes (i.e., \emph{Cogproc})}, e.g., \emph{Tentativeness}, \emph{Differentiation}, and \emph{Insight}. On the other hand, Weibo active citizens use more frequently words belonging to LIWC categores such as \emph{informal and nonflu (nonfluent)} that are similar to their correlated N-grams (see Table~\ref{t:bowweibo}).

\section{Conclusion}
In this paper, we have presented an extensive study on identifying and characterizing misinformation posters and active citizens across two different social media platforms (i.e., Twitter and Weibo) and languages (i.e. Chinese and English) for the first time. We developed a new Weibo dataset with users labeled into the two categories and repurposed an existing Twitter dataset for the task. 
Our hierarchical transformer model performs best, achieving up to 80.2 and 85.1 macro F1 score on Twitter and Weibo datasets respectively. Finally, we perform a linguistic feature analysis unveiling the major differences in language use between the two groups of users across platforms. In the future, we plan to explore cross-lingual settings for the task as well as including information from different modalities such as images~\citep{sanchez-villegas-aletras-2021-point,sanchez-villegas-etal-2021-analyzing}.

\begin{table}[!t]
\caption{Simplified Chinese LIWC features associated with Weibo misinformation posters and active citizens sorted by Pearson's correlation ($r$) between the normalized frequency and the labels ($p < .001$).}
\centering
\small
\begin{tabular}{|l|l||l|l|}
\hline
\rowcolor{dGray} \multicolumn{4}{|c|}{{\bf LIWC}} \\ \hline
\rowcolor{mGray} {\bf Posters}            & {\bf r}         & {\bf Active Citizens}           & {\bf r}       \\ \hline
social  & 0.251 &  model\_pa    & 0.338 \\ \hline
prep    & 0.208 &  informal     & 0.334 \\ \hline
health  & 0.180 &  assent       & 0.332 \\ \hline
space   & 0.170 &  progm        & 0.328 \\ \hline
you     & 0.168 &  nonflu       & 0.307 \\ \hline
female  & 0.168 &  insight      & 0.298 \\ \hline
achieve & 0.166 &  practice     & 0.259 \\ \hline
sexual  & 0.162 &  tensem       & 0.258 \\ \hline
friend  & 0.160 &  adverb       & 0.220 \\ \hline
drives  & 0.160 &  swear        & 0.209 \\ \hline            
\end{tabular}

\label{t:liwctwitter}
\end{table}

\section*{Ethics Considerations}
Our work has received ethical approval from the Ethics Committee of our department (Reference Number 025470) and complies with the Weibo and Twitter data policies for research. 

To ensure the anonymity of the data, we only share the user's ID, rather than the username that appears on the platform. We do not share the data for non-research purposes.

\section*{Acknowledgments}
We would like to thank Danae Sánchez Villegas, Mali Jin, George Chrysostomou, Xutan Peng and all the anonymous reviewers for their valuable feedback. Pu Niu is the corresponding author and supported by China Postdoctoral Science Foundation (No. 2021M701371).

\bibliographystyle{ACM-Reference-Format}
\bibliography{naive}
\end{document}